%% file: root.tex
\documentclass[letterpaper, 10 pt, conference]{ieeeconf}  

\IEEEoverridecommandlockouts                              

\usepackage{graphicx}
\usepackage{graphics} 
\usepackage{epsfig} 
\usepackage{subcaption}

\usepackage[font=small]{caption}
\usepackage{mathptmx} 
\usepackage{times} 
\usepackage{amsmath, amsfonts} 
\usepackage{amssymb}  
\usepackage{gensymb}
\usepackage{algorithm}
\usepackage{algpseudocode}
\usepackage{tabularx,booktabs}
\usepackage{xcolor}
\usepackage{sidecap}

\newcommand{\Rc}{\mathcal{R}}

\newcommand{\nv}{\mathbf{n}}

\newcommand{\yv}{\mathbf{y}}

\newcommand{\Bv}{\mathbf{B}}

\newcommand{\Dv}{\mathbf{D}}

\newcommand{\Vv}{\mathbf{V}}

\newcommand{\Xv}{\mathbf{X}}

\newcommand{\betav      }{\boldsymbol \beta      }

\newcommand{\muv        }{\boldsymbol \mu        }

\newcommand{\Gammav     }{\boldsymbol \Gamma     }

\newcommand{\Sigmav     }{\boldsymbol \Sigma     }

\title{\LARGE \bf
GUTS: Generalized Uncertainty-Aware Thompson Sampling for Multi-Agent Active Search
}

\author{Nikhil Angad Bakshi$^{1}$, Tejus Gupta$^{1}$, Ramina Ghods$^{1}$ 
and Jeff Schneider$^{1}$
\thanks{$^{1}$N. A. Bakshi, T. Gupta, R. Ghods and J. Schneider are with the Robotics Institute, School
of Computer Science, Carnegie Mellon University, Pittsburgh, PA 15213
        {\tt\small \{nabakshi, tejusg, rghods, schneide\}@cs.cmu.edu}}%
}

\begin{document}

\maketitle
\thispagestyle{empty}
\pagestyle{empty}

\begin{abstract}

Robotic solutions for quick disaster response are essential to ensure minimal loss of life, especially when the search area is too dangerous or too vast for human rescuers. We model this problem as an asynchronous multi-agent active-search task where each robot aims to efficiently seek objects of interest (OOIs) in an unknown environment. This formulation addresses the requirement that search missions should focus on quick recovery of OOIs rather than full coverage of the search region. Previous approaches fail to accurately model sensing uncertainty, account for occlusions due to foliage or terrain, or consider the requirement for heterogeneous search teams and robustness to hardware and communication failures. We present the Generalized Uncertainty-aware Thompson Sampling (GUTS) algorithm, which addresses these issues and is suitable for deployment on heterogeneous multi-robot systems for active search in large unstructured environments. We show through simulation experiments that GUTS consistently outperforms existing methods such as parallelized Thompson Sampling and exhaustive search, recovering all OOIs in $80\%$  of all runs. In contrast, existing approaches recover all OOIs in less than $40\%$ of all runs. We conduct field tests using our multi-robot system in an unstructured environment with a search area of $\approx 75,000$ $m^2$. Our system demonstrates robustness to various failure modes, achieving full recovery of OOIs (where feasible) in every field run, and significantly outperforming our baseline.

\end{abstract}

\input{introduction.tex}

\input{related_work.tex}
  
\input{contributions.tex}

\input{system_description}

\input{problem_formulation}

\input{generalized_nats.tex}

\input{experiments.tex}

\input{conclusions_future_work.tex}

\section{Acknowledgement}
This material is based upon work supported by the U.S. Army Research Office and the U.S. Army Futures Command under Contract No. W911NF-20-D-0002. Authors would like to acknowledge the contributions of Conor Igoe, Arundhati Banerjee, Herman Herman, Jesse Holdaway, Prasanna Kannappan, Luis E. Navarro-Serment, Matthew Schnur, Wennie Tabib and Vamsavardan Vemuru.

\end{document}

%% file: introduction.tex
\section{Introduction}

Multi-robot systems have been increasingly applied to large-scale search and rescue operations \cite{sar1, sar2, sar3, sar4}. 
Such emergency operations are time-critical, hence search efficiency is crucial. While human operators can remotely control a small number of vehicles, they cannot efficiently coordinate larger teams. This motivates the requirement for decentralized multi-robot systems capable of efficient active search in large unstructured terrestrial environments.



Active search is the problem of making efficient sequential data-collection decisions to identify sparsely located OOIs, while adapting to new sensing information \cite{c1, spats}. The basic framework for active search consists of maintaining a posterior probability distribution over locations of OOIs and optimizing for information-seeking actions. These actions must balance surveying unseen areas (exploration) and confirming suspected OOIs (exploitation) \cite{activeRL1, activeRL2, activeRL3}. 




  \begin{figure}[t!]
    \centering
    \includegraphics[width=0.47\textwidth]{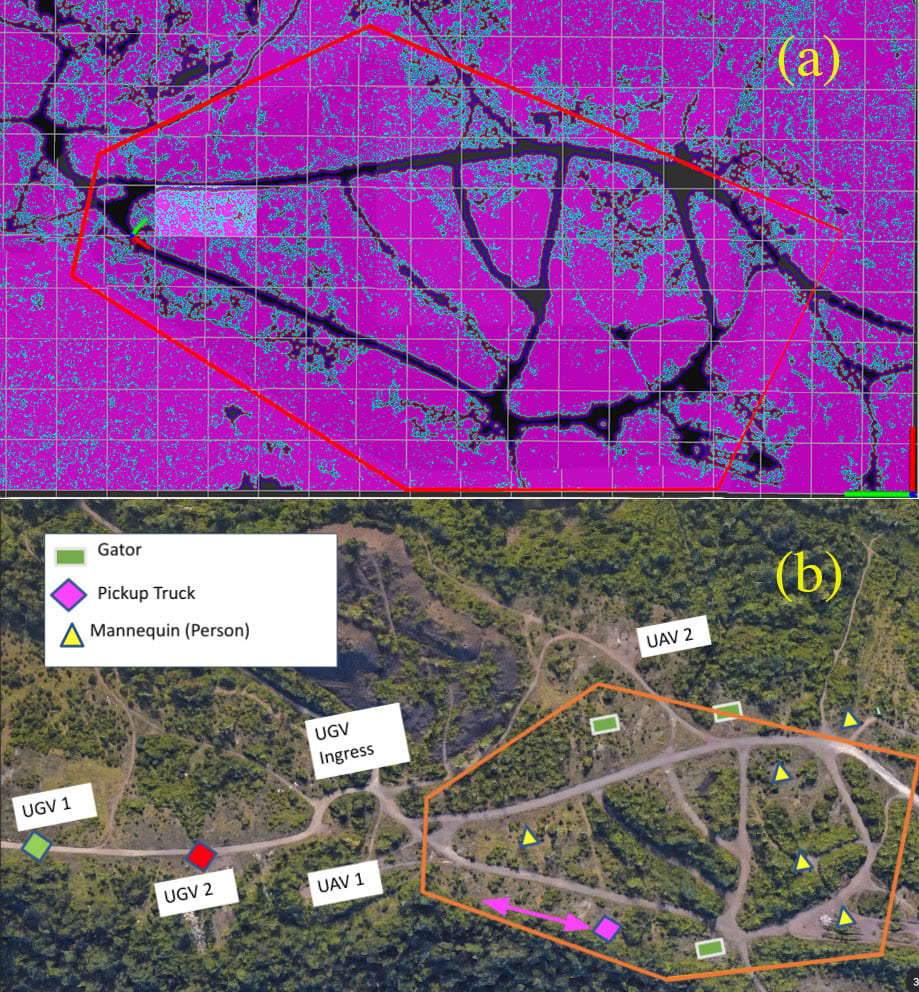}
    \caption{\footnotesize{Test-field in Pittsburgh, PA. (a) shows the traversal costmap and the search polygon (purple is high cost and black is low cost), and (b) shows an overhead image with the most common OOI locations and launching areas of the robots. The area inside the polygon is roughly 75,000 sq meters.}}
    \label{fig:global_costmap}
    \vspace*{-5mm}
\end{figure}

Even though this basic framework is straightforward, several additional challenges need to be addressed for deploying these algorithms on a multi-robot system in realistic search and rescue scenarios. OOIs may be occluded to overhead or lateral views; hence, a combination of ground and aerial search vehicles is essential to canvass such a search region thoroughly. This requires the search algorithm to reason through occlusions, viewshed constraints, as well as each robot's sensing model and navigational constraints to plan their search actions. Similarly, though multi-robot teams can theoretically partition the search region for exploration efficiency \cite{partition}, generating such a partitioning is challenging in a decentralized system. Though we would like the robots to share as much sensing information as possible for effective coordination, we cannot rely on consistent communication. Therefore, we require our system to be asynchronous and capable of making intelligent decisions with partial information. Finally, the number of OOIs in the search region is usually unknown and sensing actions are noisy \cite{noise_det1, noise_det2, noise_det3}. Previous search methods make strong simplifying assumptions \cite{ergodicbiorobotics1, swarm1} and don't deal with these challenges fully, making it intractable to deploy them on realistic search settings.

We propose a novel search algorithm, Generalized Uncertainty-aware Thompson Sampling (GUTS) that follows the parallelized Thompson Sampling framework~\cite{c1, spats, parallel_ts_kandasamy}. We propose an improved reward function, realistically model sensing noise in UAVs and UGVs, and computationally optimize it to be run on large environments. 

We design GUTS to be applicable to any unstructured search region, without assumptions on number of OOIs, and to be robust to communication and hardware failures. For instance, our search team continues to operate even if some robots fail and stop responding. 

We demonstrate our search system consisting of UGVs and UAVs on field tests in a large unstructured natural environment in Pittsburgh, PA shown in Fig.~\ref{fig:global_costmap} and on simulated tests. Our results show the robustness of our system, along with superior search performance compared to existing approaches. To the best of our knowledge, this is the first heterogeneous multi-robot system for asynchronous multi-agent active search built at this scale that is robust to communication and robot failures and operates without any human direction or manual subdivision of the search region. 

%% file: related_work.tex
\section{Related Work}
 \label{sec:related_work}

Previous works on multi-robot search systems work in more specialized settings (e.g., homogeneous teams, single agent search) and make stronger assumptions (perfect communication, homoscedastic noise model). For instance, coverage-based search methods \cite{exhaust1, exhaust2, exhaust3, gtsp} exhaustively navigate the search space. They assume that the search area is mapped and do not adapt to new information online. A similar criticism exists for frontier-based methods like \cite{frontier1}. 

Bayesian active-search methods \cite{noise_det1, noise_det2} \cite{noise_det3} can accurately model observation uncertainty when searching for sparse signals with a single agent. However, the deterministic nature of their search policies makes extending these methods to the multi-agent setting difficult without centralization. Recent work has modelled the active-search problem as a POMDP and trained search policies in simulation using reinforcement learning \cite{activeRL2, activeRL1, activeRL3, conor, decpomdp}. It is usually difficult to deploy these methods on robotic systems because of their high sample complexity and unrealistic simulations. 

Ghassemi et al. \cite{swarm1} employ an information-theoretic approach to asynchronous multi-agent active search, but work with a homogenous swarm assuming perfect communication and a homoscedastic noise model. Salman et al. \cite{ergodicbiorobotics1} generate trajectories such that the time spent in an area corresponds to the likelihood of the presence of OOIs. However, their method does not incorporate new observations and relies on perfect communication and a strong unified prior on locations of OOIs. 

Dec-MCTS \cite{treemcts} is a multi-agent active-perception algorithm that has shown promising simulated results. Dec-MCTS is a decentralized version of MCTS where all robots optimize their trajectories asynchronously but achieve coordination by sharing their partial search trees and do not model uncertainties in realistic perception pipelines.

Noise-Aware Thompson Sampling (NATS) \cite{c1} is an asynchronous multi-agent active-search algorithm that can deal with heteroscedastic noise models and has shown promising simulated results. NATS uses Myopic Posterior Sampling~(MPS)~\cite{mps_kandasamy} to ensure that the agents take diverse sensing actions without centralization. 
Thompson Sampling~(i.e. MPS) \cite{tsoriginal} is an online optimization method that balances exploration and exploitation by maximizing the expected reward assuming that a sample from the posterior is the true state of the world \cite{tstutorial}. This stochastic nature is the key attribute that makes it an excellent candidate for an asynchronous multi-agent setting with unreliable communication as it promotes some diversity in action selection. 

%% file: contributions.tex
\section{Contributions}
\label{sec:contrib}
\begin{itemize}
    \item We propose the Generalised Uncertainty-aware Thompson Sampling (GUTS) algorithm, that uses an improved reward design, more accurately models sensing noise, and is computationally optimized over prior parallelized Thompson Sampling methods \cite{c1, spats}.
    \item We present a working system of heterogeneous robots that is capable of asynchronous multi-agent active search in large unstructured natural environments without a central planner. We design our multi-agent search algorithm to be robust to loss of robots mid-search, noisy observations and communication breakdown without any human intervention.
    \item We experimentally show superior search success rate and search efficiency compared to existing search methods, namely NATS and coverage-based search, in field and simulated experiments.
\end{itemize}

%% file: system_description.tex
\section{System Description}
\label{sec:sys_des}
\begin{figure}[t!]
    \centering
     \begin{subfigure}{0.20\textwidth}
         \centering\includegraphics[width=\textwidth]{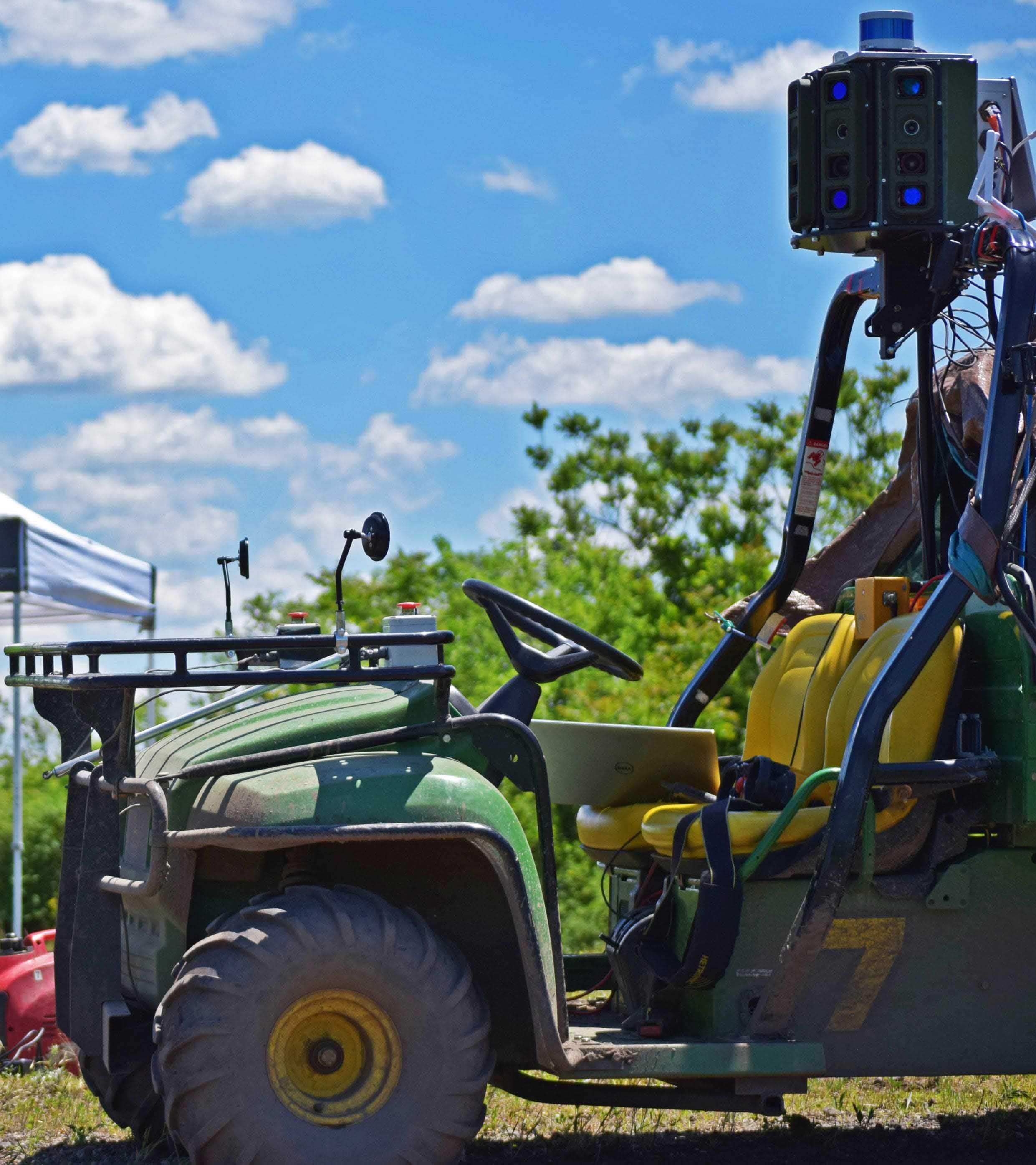}
     \end{subfigure}
     \hfil
     \begin{subfigure}{0.215\textwidth}
         \centering\includegraphics[width=\textwidth]{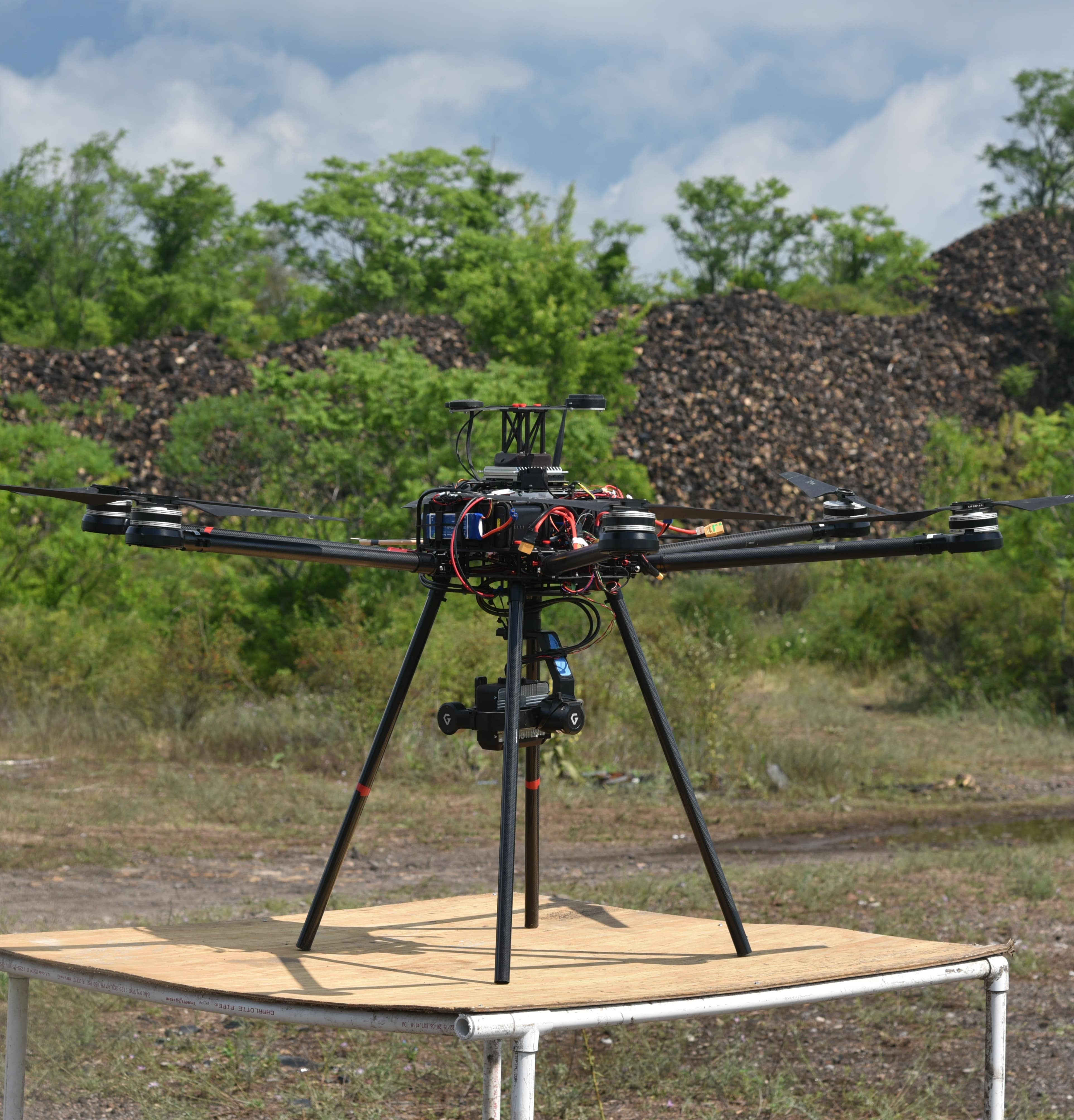}
     \end{subfigure}
\caption{Left: RecBot UGV. Right: Lil Hexy UAV.}
\label{fig:gator_hexy}
\vspace*{-5mm}

\end{figure}

This section describes our multi-robot system on which we evaluate our search algorithm.
For the UGV, we use the RecBot, a John Deere E-Gator
utility vehicle that has been retro-fitted with a drive-by-wire
autonomy kit (Fig.~\ref{fig:gator_hexy}-left). For the UAV, we use the Lil Hexy \footnote{https://vscl.tamu.edu/vehicles/little-hexy/} air-frame paired with a PixHawk flight controller \footnote{https://pixhawk.org/} as our base platform (Fig.~\ref{fig:gator_hexy}-right). We added a custom camera payload and additional onboard compute, namely, NVIDIA Jetson AGX Xavier to this platform.

We summarize the components of our autonomy stack necessary to understand the work in this paper below.

\begin{figure}[ht!]
    \centering
         \begin{subfigure}{0.25\textwidth}
         \centering\includegraphics[width=\textwidth]{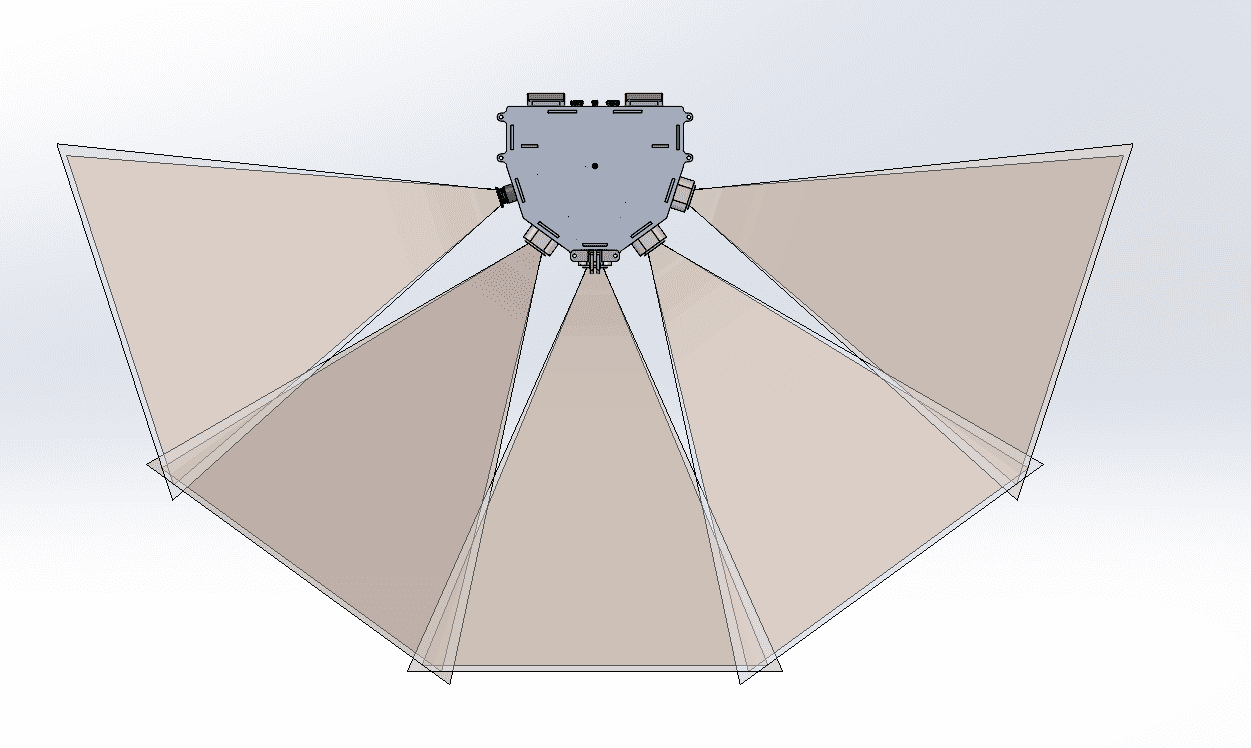}
     \end{subfigure}
     \hfil
     \begin{subfigure}{0.22\textwidth}
         \centering\includegraphics[width=\textwidth]{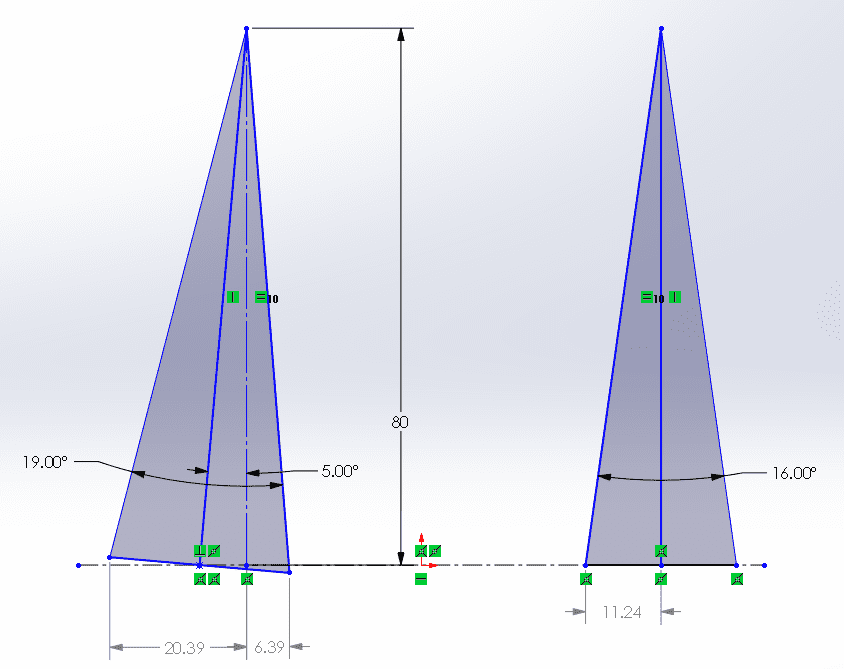}
     \end{subfigure}

    \caption{Left: Top-down view of UGV camera field-of-view. Span is 192\degree  horizontally and 37\degree  vertically. Right: UAV camera field-of-view. Span is $19\degree   \times 16 \degree$. At a flight height of 80m, this leads to approximately $27.8m \times 22.5m$ rectangle of visibility on the ground.}
    \label{fig:ugv_fov}
    \vspace*{-5mm}
\end{figure}

\begin{figure}[ht!]
     \centering
     \begin{subfigure}{0.20\textwidth}
         \centering
         \includegraphics[width=\textwidth]{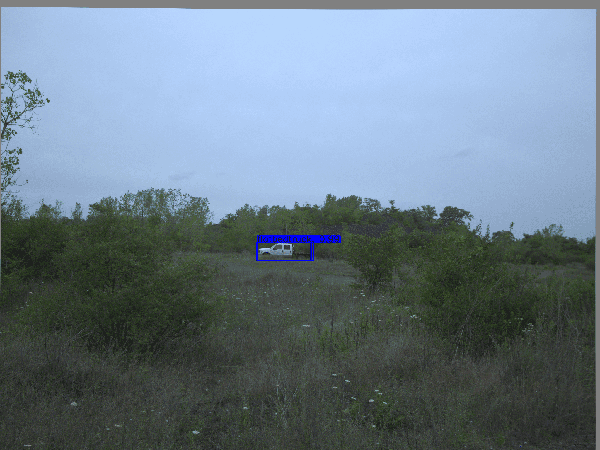}
         \caption{Pickup Truck}
         
     \end{subfigure}
    \begin{subfigure}{0.20\textwidth}
         \centering
         \includegraphics[width=\textwidth]{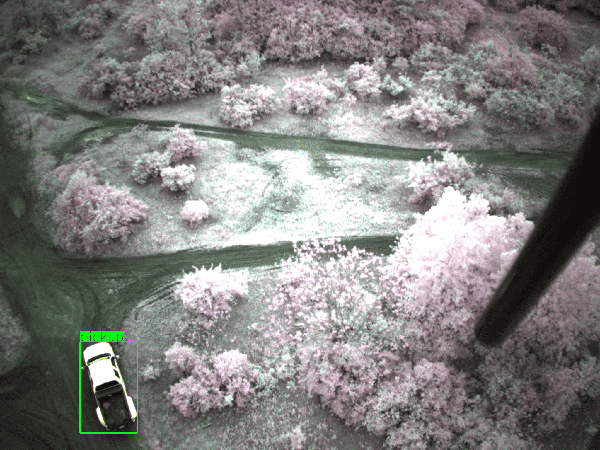}
         \caption{Pickup Truck}
     \end{subfigure}
     \begin{subfigure}{0.20\textwidth}
         \centering
         \includegraphics[width=\textwidth]{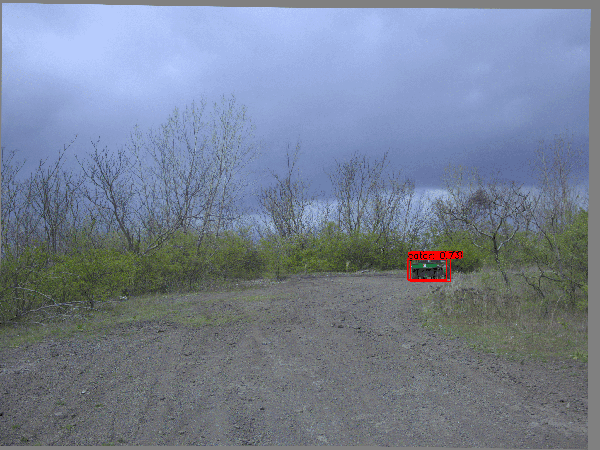}
         \caption{E-Gator}
     \end{subfigure}
     \begin{subfigure}{0.20\textwidth}
         \centering
         \includegraphics[width=\textwidth]{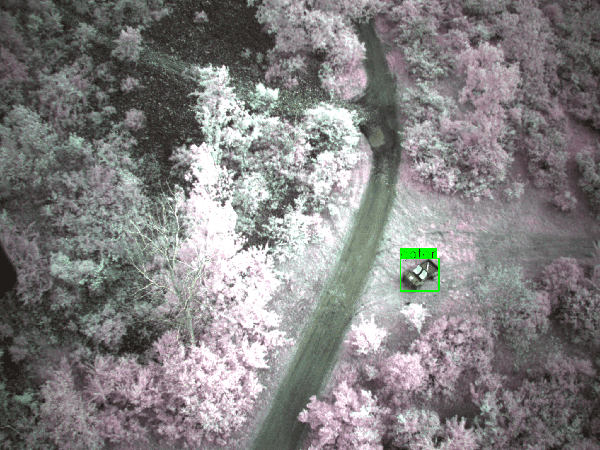}
         \caption{E-Gator}
     \end{subfigure}
     \begin{subfigure}{0.20\textwidth}
         \centering
         \includegraphics[width=\textwidth]{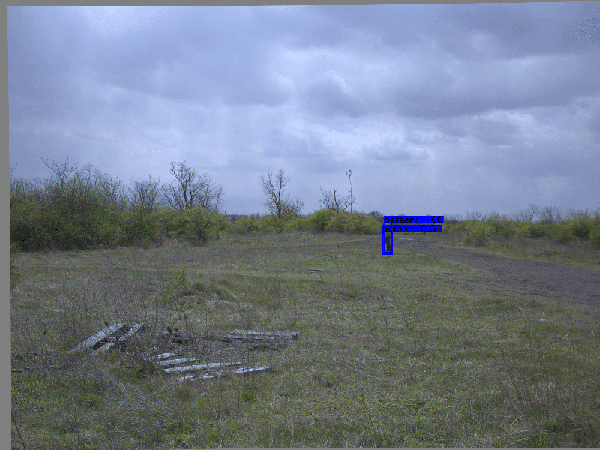}
         \caption{Person}
         
     \end{subfigure}
     \begin{subfigure}{0.20\textwidth}
         \centering
         \includegraphics[width=\textwidth]{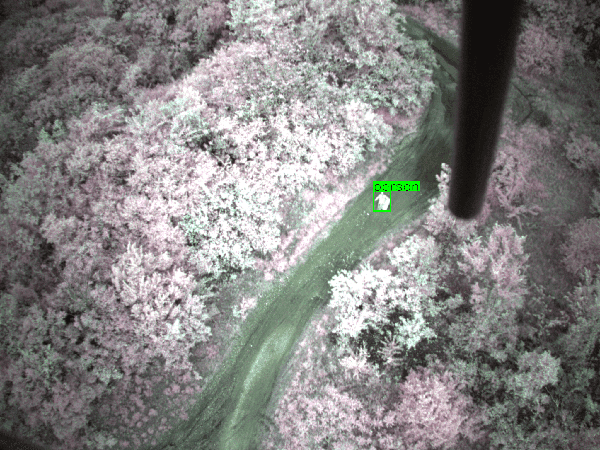}
         \caption{Person}
     \end{subfigure}
    \caption{Object detector: qualitative results. We show images and OOI detections for the UGV (left) and the UAV (right)}
    \label{fig:det_quali}
    \vspace*{-5mm}
\end{figure}

\begin{itemize}
    \item \textbf{Camera FOV:} The sensor pod on the UGV has a $37\degree$ vertical and $192\degree$ horizontal FOV (Fig.~\ref{fig:ugv_fov}-left). The sensor module consists of five RGB cameras, each of 12 MP, to allow for accurate OOI detection up to several hundred meters away. A sensor pod with one RGB camera of 5 MP resolution is mounted on the UAV pointed downward with a $5\degree$ angle with the vertical, tilted in the direction of the travel of the UAV~(Fig.~\ref{fig:ugv_fov}-right). The $19\degree \times 16\degree$ FOV on this camera would cover an area of $27.8m \times 22.5m$ from a flight height of 80m.
    \item \textbf{Object Detector:} For the UGV, we use a customized version of the widely used YOLOv3 neural network architecture \cite{yolov3} with fast inference on the NVIDIA RTX 6000 GPU given very high-resolution input images. For the UAV detector, the computing power is limited; so we use a less computationally intensive model \cite{huangdetector} and run it on the NVIDIA Jetson AGX Xavier.
    \item \textbf{Object Tracking:} The perception module uses a Kalman filter \cite{kalman} to track possible OOIs. We describe our sensing model in detail in section \ref{sec:nats_ext}. For example, the estimated OOI locations detected by the ground-vehicles have more uncertainty along the range to the object compared to the bearing direction. This is represented by the covariance associated with a track returned by the Kalman filter. Moreover, this uncertainty increases with object distance from the observer. 

    \item \textbf{Navigation for UGV:}
    We utlise the SBPL planner\footnote{https://github.com/sbpl/sbpl} with a local lookahead controller to chart safe paths to waypoints provided by the search module being presented in this paper. We update the offline global map with on-robot sensing to generate the traversal costmap.
    \item \textbf{Objects of Interest:} The OOIs are people (mannequins), a John Deere E-Gator, or a pickup truck. Fig. \ref{fig:det_quali} shows example frames from the UGV and UAV camera systems with successful detections.
\end{itemize}

\vspace{-0.05cm}

%% file: problem_formulation.tex
\section{Problem Formulation}

\label{sec:prob_form}

We model the search region as a grid with a cell size of 30m x 30m. We have access to the costmap for ground robot traversability. Fig.~\ref{fig:global_costmap} shows an overhead view and the costmap for the field testing site in Pittsburgh, PA. This costmap was generated from a drone flyover several months before field testing. Some parts of the map have changed since then, but our on-robot sensing and mapping system can recognize changes like blocked paths. Due to the ubiquity of satellite imagery, it is reasonable to assume such approximate map information of the search region to be available. 

As is evident from Fig. \ref{fig:global_costmap}, several areas cannot be accessed or viewed by ground vehicles. In addition, large portions of the map are covered with dense foliage, which may obscure any OOIs hidden under it from the view of the air vehicles. This necessitates heterogenous coordination to explore the search region completely.

We model the active search problem as follows:
\begin{itemize}
    \item We give the same search region to all robots. 
    (see the red polygon in Fig. \ref{fig:global_costmap} for an example).
    \item The OOIs are sparsely placed and static. They need to be located as fast as possible with high certainty.
    \item Each robot must plan its next data collection action on-board, i.e., no central planner exists.
    \item The robots may communicate their locations and observations with each other; however, the algorithm should function anywhere in the spectrum of total absence of communication to perfect communication.
\end{itemize}
We note that the objective is \textit{not to cover} the entire search space. Rather it is specifically to locate all OOIs as fast as possible. 

More formally, we represent the locations of OOIs in our 2D grid representation using a sparse matrix $\Bv \in \mathbb{R}^{M_1 \times M_2}$. Let $\betav \in \mathbb{R}^{M}$ be the flattened version of matrix $\Bv$, where $M = M_1M_2$. This vector is sparsely populated, with $1$ at locations corresponding to OOIs and $0$ elsewhere. We model the uncertainty in our observations by
\begin{equation}
\label{eqn:sensingmodel}
 \yv_t= \textrm{clip}(\Xv_t\betav \pm \nv_t,0,1) \textrm{ with } \nv_t\sim\mathcal{N}^{+}(0, \Sigma_t)   
\end{equation}

where $\Xv_t \in \mathbb{R}^{Q\times M}$ describes the sensing matrix such that each row in $\Xv_t$ is a one-hot vector indicating one of the grid cells in view of the robot at timestep $t$, and $Q$ is the total number of grid cells the robot can view. $\yv_t \in \mathbb{R}^{Q \times 1}$ is the resultant observation including the additive depth-aware noise vector $\nv_t \in \mathbb{R}^{Q \times 1}$. This depth-aware noise encodes the intuition that observation uncertainty increases with distance to the robots, and we model it by having the diagonal elements of the noise covariance matrix $\Sigma_t$ monotonically increase with the distance of the observed cell from the robot. The off-diagonal elements of $\Sigma_t$ are zero. The noise is sampled from a positive half-Gaussian distribution $\mathcal{N}^{+}(0, \Sigma_t)$ and is added for cells without OOIs and subtracted for cells with OOIs. The observation $y_t$ is clipped to be within 0 and 1.

Let $\Dv^j_t$ be the set of observations available to robot $j$ at timestep $t$. $\Dv^j_t$ comprises of $(\Xv_t, \yv_t)$ pairs collected by robot $j$ as well as those communicated to robot $j$ by other robots. Let the total number of sensing actions by all agents be $T$. Our main objective is to sequentially optimize the next sensing action $\Xv_{t+1}$ based on $\Dv^j_t$ at each timestep $t$ to recover the sparse signal $\betav$ with as few measurements $T$ as possible. Each robot optimizes this objective based on its own partial dataset $\Dv^j_t$ in a decentralized manner.

%% file: generalized_nats.tex
\section{Generalised Uncertainty-aware Thompson Sampling (GUTS)}
\label{sec:nats_ext}

This section presents the GUTS algorithm. The framework for decision-making follows \cite{c1}. 
Each robot $j$ asynchronously estimates the posterior distribution over OOI locations based on its partial dataset $\Dv_t^j$. During the action selection stage, each robot generates a sample from this posterior and optimizes a reward function for this sampled set of OOI locations. Sec. \ref{sec:nats_posterior} and Sec. \ref{sec:nats_reward} describe both these steps for the Noise-Aware Thompson Sampling (NATS) algorithm presented in \cite{c1}. We utilize the posterior computation step in Sec. \ref{sec:nats_posterior} for GUTS.

In Sec. \ref{sec:guts_reward} and onward, we describe the new reward function to improve search performance, the depth-aware noise models for our UGVs and UAVs, and speed-up techniques to optimise the code to run on large environments.


\subsection{Calculating Posterior}
\label{sec:nats_posterior}

Each robot assumes a zero-mean gaussian prior per entry of the vector $\betav$ s.t. $p_0(\beta_m)=\mathcal{N}(0,\gamma_m)$. The variances $\Gammav = diag([\gamma_1 ...\gamma_M])$ are hidden variables which are estimated using data. We follow \cite{sbl, inverse_gamma} and use a conjugate inverse gamma prior on $\gamma_m$ to enforce sparsity s.t. $p(\gamma_m)=IG(a_m,b_m) = \frac{b_m^{a_m}}{\Gamma(a_m)}\gamma_m^{(-a_m-1)}e^{-(b_m/\gamma_m)}\: \forall \: m \in \{1...M\}$. We estimate the posterior distribution on $\betav$ given data $\Dv_t^j$ for robot $j$ using Expectation Maximisation~\cite{em}.

We can write analytic expressions for the E-step (estimating $p(\betav|\Dv_t^j,\Gammav) = \mathcal{N}(\muv, \Vv)$) and M-step (computing $\max_{\Gammav} p(\Dv_t^j| \betav,\Gammav)$) respectively:
\begin{equation}
\label{eqn:e_step}
    \Vv = (\Gamma^{-1}+\Xv^T \Sigmav \Xv)^{-1}; \muv = \Vv\Xv^T\Sigmav \yv
\end{equation}
\vspace{-0.5cm}
\begin{equation}
\label{eqn:m_step}
    \gamma_m = ([\Vv]_{mm} + [\muv]^2_m + 2b_m)/(1+ 2a_m)
\end{equation}
where $\Xv$ and $\yv$ are created by vertically stacking all measurements $(\Xv_t,\yv_t)$ in $\Dv_t^j$, and $\Sigmav$ is a diagonal matrix composed of their corresponding depth-aware noise variances.

Each robot estimates $p(\betav |\Dv_t^j) = \mathcal{N}(\muv,\Vv)$ on-board using its partial dataset $D_t^j$. We drop the index $j$ from equations \eqref{eqn:e_step} and \eqref{eqn:m_step} for clarity. We set the values of $a_m=0.1$ and $b_m=1$ as these were found to be effective in \cite{c1}. Finally, agent $j$ samples from the posterior $\tilde{\betav} \sim p(\betav|\Dv^j_t)$.

\subsection{Choosing Next Sensing Action}
\label{sec:nats_reward}

Each robot chooses the next sensing action $\Xv_{t+1}$ by assuming that the sampled set of OOI locations $\tilde{\betav}$ is correct. Specifically, let $\hat{\betav} (\Dv^j_t \cup (\Xv_{t+1}, \yv_{t+1}))$ be our expected estimate of the parameter $\beta$ using all available measurements $\Dv^j_t$ and the next candidate measurement $(\Xv_{t+1}, \yv_{t+1})$. Then the reward function to evaluate each possible sensing action is defined as:
\begin{equation}
    \label{eqn:reward_orig}
    \Rc(\tilde{\betav}, \Dv^j_t, \Xv_{t+1}) = -\mathbb{E}_{\yv_{t+1}|\Xv_{t+1},\tilde{\betav}}[||\tilde{\betav} - \hat{\betav} (\Dv^j_t \cup (\Xv_{t+1}, \yv_{t+1}))||^2_2]
\end{equation}

We select the sensing action $\Xv_{t+1}$ that maximizes this reward. The reward function is stochastic due to the sampling of $\tilde{\beta}$ and this ensures that the search actions selected by the robots are diverse.

\subsection{GUTS Reward function}
\label{sec:guts_reward}

During action-selection, each robot optimizes a sensing location by maximizing the reward function in equation (\ref{eqn:reward_orig}). We observed that this reward function leads to poor search performance at the start of the run. At the start of the episode, the posterior belief over OOI locations is very uncertain and all possible next waypoints have almost-zero probability of finding an object. Hence, this reward function leads to excessively explorative behavior.

We modify the reward function to 
\vspace{-7mm}

\begin{multline}
\label{eqn:reward_simple}
\Rc(\tilde{\betav}, \Dv^j_t, \Xv_{t+1}) = -\mathbb{E}_{\yv_{t+1}|\Xv_{t+1},\tilde{\betav}}[||\tilde{\betav} - \hat{\betav} (\Dv^j_t \cup (\Xv_{t+1}, \yv_{t+1}))||^2_2] \\
- \lambda \times I(\tilde{\betav}, \hat{\betav})
\end{multline}

\vspace{-3mm}

Where $\lambda(= 0.01)$ is a hyperparameter that reduces the reward for a search location if the estimated $\hat{\betav}$ does not have high likelihood entries in common with the sample at the current step $\tilde{\betav}$. This change improves the discriminatory power of the reward and encourages exploitative actions early in a search run. Formally, let $\hat{k}$ and $\tilde{k}$ be the number of non-zero entries in $\hat{\betav}$ and $\tilde{\betav}$, then the indicator function $I(.)$ is defined as:
\begin{equation*}
    I(\tilde{\betav}, \hat{\betav)} = \begin{cases}
    0 \parbox[t]{.35\textwidth}{, if any matches between top $\frac{\hat{k}}{2}$ entries in $\hat{\betav}$ \:\:\:\: and top $\frac{\tilde{k}}{2}$ entries in $\tilde{\betav}$} ,\\
    1 \text{, otherwise}
    \end{cases}
\end{equation*}


Our experiments show that this modified reward function achieves better search efficiency.

\subsection{UAV Sensing Action Model} The flight height of each UAV is fixed between $70m-90m$, with an approximate FOV of $30m \times 30m$ on the ground below it. We consider flight paths that are straight lines and and integrate over cells that will be flown over en-route to the final waypoint in the reward computation.


\subsection{UGV Sensing Action model}
We model the FOV of the UGV as a $30m \times 60m$ space directly in front of it. In the GUTS grid representation, this would be the two cells along the bearing of the UGV.
%
%
We approximate the FOV conservatively as opposed to the full FOV depicted in Fig.~\ref{fig:ugv_fov}. Overestimating the FOV would be disastrous since GUTS would then erroneously assume some parts of search region have been searched and never find OOIs in that region. Note that the perception module still reports positive detections even beyond the modelled FOV, hence there is no loss of information.

We do not consider possible OOIs en route to a candidate sensing action $\Xv_t$ for computing the reward function for UGVs because computing and integrating over trajectories for each possible sensing action $\Xv_t$ is expensive. Each robot still logs and shares observations along the way to a waypoint even though they weren't explicitly accounted for. The problem of integrating trajectory-level information gain for UGVs will be tackled in future work.

\subsection{Modified Depth-Aware Noise Modelling}

We describe our heteroscedastic noise-model for OOI observations in this section. NATS \cite{c1} models this noise using a diagonal covariance matrix where the diagonal elements are proportional to the distance between the observed grid cell and the robot. Intuitively, this represents our confidence in existence of a detected OOI at the grid cell in consideration. However, this is not a quantity that is actually available for learning-based perception systems.

We combine the detection confidence of our object detector with the location uncertainty obtained through the Kalman Filter to estimate this confidence. We employ the following heuristic to approximate the existence uncertainty for positive observations:
\begin{equation*}
    \sigma^2 = min\left(0.5, \frac{\text{Vol. of OOI loc uncertainty ellipsoid }(m^3)}{\text{OOI detection confidence} \times 1000}\right) 
\end{equation*}

This formula is based on the intuition that the location uncertainty obtained by the Kalman filter implicitly encodes existence uncertainty as well. We re-use the original depth-aware noise-model used in \cite{c1} for negative observations.

\subsection{Accelerating GUTS}
\label{sec:acc_guts}
We describe a simple example to make the scale of the speedup clear. For an area of 0.7 sq.km, the original NATS implementation \cite{c1} took over 60 minutes to compute the first waypoint. After the following three changes, we cut the waypoint selection time to under two minutes:
\begin{itemize}
    \item GUTS needs to compute $\Vv$ in \eqref{eqn:e_step} repeatedly for candidate sensing action $\Xv_{t+1}$. The matrix inversion step is expensive. We leverage the diagonal nature of $(\Gammav^{-1}+\Xv^T \Sigmav \Xv)^{-1}$ to speed up inversion by simply inverting the diagonal elements. 
    \item Additionally, the $\Xv^T\Sigma\Xv$ term in \eqref{eqn:e_step} can be calculated additively as follows:
        \begin{equation*}
        \begin{bmatrix}\Xv_1\\ \Xv_2\end{bmatrix}^T\begin{bmatrix}\Sigma_1 & 0\\0 & \Sigma_2\end{bmatrix} \begin{bmatrix}\Xv_1\\ \Xv_2\end{bmatrix}= \Xv_1^T\Sigma_1\Xv_1 + \Xv_2^T\Sigma_2\Xv_2
    \end{equation*}
    This leverages the independence of sensing noise in different grid cells and results in significant-speedup later in the run when $\Xv$ is high-dimensional.
    \item For $\muv$ in Eqn.~\ref{eqn:e_step}, we exploit the sparse nature of the sensing matrix $X$ to speed up computation \cite{scipy}.
\end{itemize}

Since compute is limited on the UAVs, we have an additional sampling parameter that subsamples the possible sensing actions the drone could take to dial in the planning time as needed. Typically we tested within the 1\% to 10\% subsampling range.

%% file: experiments.tex
\section{Experiments and Results}
\label{sec:experiments}

\begin{table*}[ht!]
\vspace*{-2mm}
\caption{Field Testing Results: We report the search parameters (team size, total search area, total OOIs, algorithm used) with evaluation metrics (runtime, OOIs found, and search efficiency T/C) for each run. We observe that GUTS outperforms the coverage-based planner in terms of search efficiency (T/C). We also note runs with communication breakdown (CB), hardware failures (HF), and issues with detectors (ND = noisy detector).}

\label{table_field_results}
\begin{tabularx}{\textwidth}{@{}l*{10}{c}c@{}}
\toprule
Run & Team Size (J) & Search Area ($m^2$) & Runtime (s) & OOIs found (C) &  Total OOIs (M) &$T$ & $T/C$  &Algorithm & Notes\\
\midrule
1 & 1 UAV & $\sim75,000$ & 1600 & 4 & 7 & 27 & 6.75  &Coverage & -\\
2 & 1 UAV & $\sim75,000$ & 1500 & 5 & 8 & 28 & 5.6  &Coverage & -\\
3 & 1 UAV & $\sim75,000$ & 1000 & 2 & 8 & 8 & 4  & 5\% GUTS & -\\
4 & 1 UAV & $\sim75,000$ & 1600 & 4 & 8 & 16 & 4  & 5\% GUTS & -\\
5 & 1 UAV & $\sim75,000$ & 500 & 3 & 4 & 5 & 1.67  & GUTS & -\\
6 & 2 UGVs & $\sim75,000$ & 1000 (avg) & 3 & 3 & 3 & 1  & GUTS& ND\\
7 & 2 UGVs & $\sim75,000$ & 600 (avg) & 6 & 6 & 6 & 1  & GUTS& CB, ND\\
8 & 2 UGVs 1 UAV & $\sim75,000$ & 1200 (avg) & 6 & 6 & 13 & 2.167 & GUTS & CB, HF, ND\\
\bottomrule
\end{tabularx}
\vspace*{-5mm}
\end{table*}

\begin{figure*}[ht!]
\includegraphics[width=.36\textwidth]{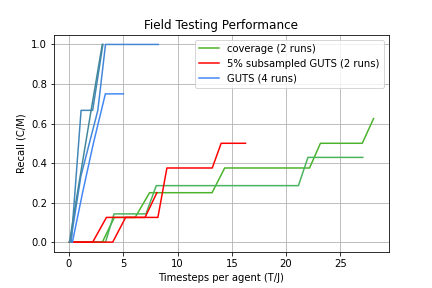}
\includegraphics[width=.32\textwidth]{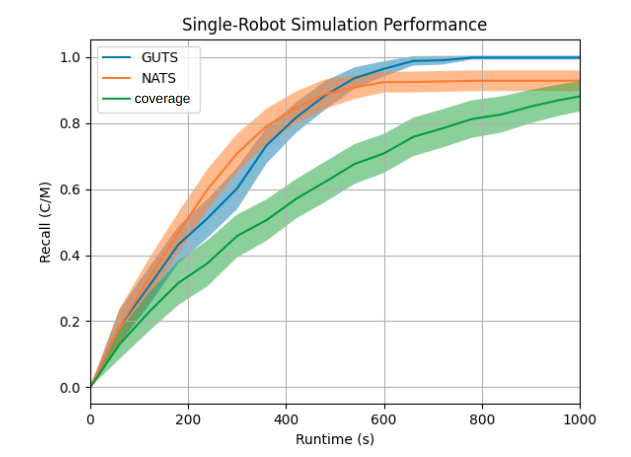}
\includegraphics[width=.32\textwidth]{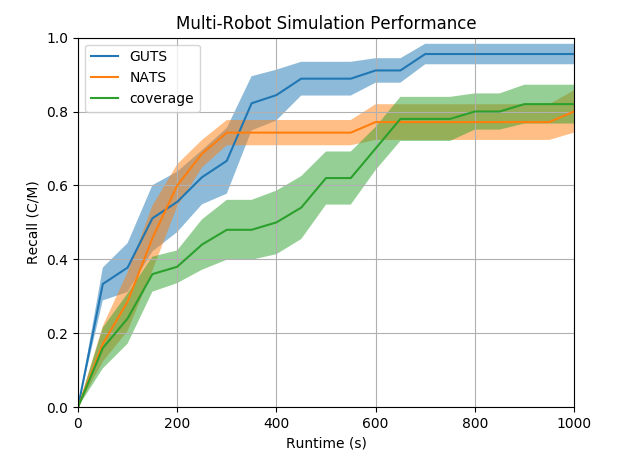}
\caption{Comparison of Search Efficiency. (Left) We visualize the search efficiency for each run in Table I. We plot the recall versus the number of sensing actions taken per robot. This graph shows that GUTS outperforms the  coverage baseline on our system. We also observe that subsampling the total set of waypoints achieves a good balance of computational efficiency and performance, and slighly outperforms the coverage baselines on UAVs.
(Middle and Right) We compare the recall rate of different search algorithms vs runtime in a single-robot and multi-robot simulation. We again observe that GUTS outperforms NATS and coverage-based search.}
\label{fig:result_graph}
\vspace*{-6mm}

\end{figure*}

Our experiments aim to demonstrate the improved search efficiency of our proposed algorithm GUTS compared to existing search methods: NATS and coverage-based search. We run a computationally optimised (similar to Sec. \ref{sec:acc_guts}) version of NATS for our experiments since the original implementation \cite{c1} is too slow for the scale of our experiments. The coverage-based baseline myopically chooses the next waypoint in an unvisited part of the search region. We evaluate these methods on a realistic multi-robot Gazebo \cite{rosgazebo} simulation. Our simulation uses a pre-mapped costmap of our field test site and accurately simulates the robots' sensing, trajectory planning, and traversal. 

We also seek to validate that GUTS can run in real-time on a heterogenous multi-robot team and can function robustly in the face of communication breakdown, hardware failures, and noisy detections. We conduct field tests with a combination of ground and aerial vehicles in a natural unstructured environment with a search area of $\sim75,000 m^2$.

\subsection{Testing Setup}
Fig.~\ref{fig:global_costmap}b shows the launch locations for the UGVs and UAVs, the search region for each robot, and the most common OOI locations. We start evaluating each robot's search performance once it enters the search polygon.

\subsection{Evaulation Metric}

Our primary evaluation metric is the search success rate, which is defined as the fraction of runs in which the search method locates all OOIs within the prescribed time budget.

For our simulated experiments, we measure search efficiency by plotting recall versus runtime for each algorithm. For our field tests, we compute the number of sensing actions per OOI found. We prefer to isolate each search algorithm's performance from physical factors such as terrain conditions in field tests and measure the recall in terms of the number of decisions rather than the total time taken. That said, we also report wall clock time for each run, and our total search time per run is always under 30 mins.

\subsection{Simulated Results}

We compare the search efficiency of GUTS with NATS and coverage-based search in Fig.~\ref{fig:result_graph}. We plot results for a single UGV (center plot) and two UGVs (right plot) with five OOIs and a 1000 second ($\approx 17$mins) runtime budget averaged across 20 runs and 10 runs, respectively.

We can see that GUTS clearly outperforms our baselines: NATS and coverage-based search, and is the only search algorithm to consistently recover all the OOIs within our time budget. GUTS has a success rate of 80\% in the multi-robot setting (i.e. it recovers all OOIs in 80\% of the runs), compared to 40\% and 30\% success rate for NATS and coverage-based search respectively.

GUTS outperforms NATS due to the modification in the reward function described in \ref{sec:guts_reward} where confirming suspected OOIs is prioritized over exploratory behaviour. This leads to a slower start but GUTS overtakes NATS towards the end of the run. Coverage-based search performs poorly because it navigates the search region exhaustively rather than optimizing its sensing actions for OOI identification.
\vspace{-1mm}

\subsection{Field-Testing Results}

We report detailed run parameters and evaluation metrics for our field tests in Table \ref{table_field_results}. We report results for 6 runs of GUTS and 2 runs of our coverage-based baseline. We plot the number of sensing actions required per agent (T/J) against the recall for each run in Fig.~\ref{fig:result_graph}-left. We observe that the GUTS algorithm shows markedly more efficient search behaviour compared to our coverage based baseline.

We also observe that 5\% GUTS, which is run on a 5\% random subsample of all possible sensing actions, slightly outperforms coverage as well. We note the runs where we observed communication breakdown or hardware failure show no degradation in performance. The search team sometimes fails to recover all the OOIs in runs with only UAVs, because some OOIs are only recoverable from the ground-vehicles. This demonstrates the need to use both UAVs and UGVs to thoroughly canvass the search region.

%% file: conclusions_future_work.tex
\section{Conclusion}
\label{sec:fw}

We present a novel algorithm, Generalized Uncertainty-aware Thompson Sampling (GUTS), for efficient multi-agent active search. GUTS uses an improved reward function, realistically models sensing noise, and is computationally optimized to be run on large environments. We show that GUTS outperforms state-of-the-art methods to more quickly and consistently recover all objects in our simulated experiments. We also validated the search efficiency and robustness of GUTS through field tests in a large unstructured environment subject to hardware and communication failures. 

Our future work would focus on modelling dynamic targets, as well as integrating path-level information-gain and topographical information for selecting sensing actions to improve search performance further for the UGVs.


We believe that greater adoption of multi-robot teams in dangerous large-scale search and rescue missions has the potential to save human lives and effort. We hope our work encourages research in more realistic search settings and quicker adoption in real search missions.



